\documentclass[10pt,twocolumn,letterpaper]{article}

\usepackage[pagenumbers]{cvpr} 

\definecolor{cvprblue}{rgb}{0.21,0.49,0.74}
\usepackage[pagebackref,breaklinks,colorlinks,allcolors=cvprblue]{hyperref}
\usepackage[most]{tcolorbox}
\tcbset{enhanced, colback=black!2, colframe=black!40, boxrule=0.3pt, arc=1mm, left=6pt, right=6pt, top=4pt, bottom=4pt}

\newtcolorbox{defbox}[1]{title=\textbf{Definition} (#1)}

\usepackage{graphicx}
\usepackage{adjustbox}
\usepackage{cuted}  
\usepackage{caption} 
\usepackage{amsmath}
\usepackage{subcaption}

\def\paperID{11684} 
\def\confName{CVPR}
\def\confYear{2026}

\title{Chain-of-Image Generation:\\ Toward Monitorable and Controllable Image Generation}

\author{
Young Kyung Kim$^{1,2}$,
Oded Schlesinger$^{1,2,*}$,
Yuzhou Zhao$^{1,2,*}$,\\
J. Mat\'ias Di Martino$^{1,3}$,
Guillermo Sapiro$^{2,4}$\\
[2mm]
$^1$~Duke University  \quad $^2$~Princeton University \quad $^3$ Universidad Católica del Uruguay \quad $^4$~Apple \\
$^*$~These authors contributed equally to this work.
}

\begin{document}
\maketitle

  

\begin{abstract}

While state-of-the-art image generation models achieve remarkable visual quality, their internal generative processes remain a ``black box.'' This opacity limits human observation and intervention, and poses a barrier to ensuring model reliability, safety, and control. Furthermore, their non-human-like workflows make them difficult for human observers to interpret. To address this, we introduce the Chain-of-Image Generation (CoIG) framework, which reframes image generation as a sequential, semantic process analogous to how humans create art. Similar to the advantages in monitorability and performance that Chain-of-Thought (CoT) brought to large language models (LLMs), CoIG can produce equivalent benefits in text-to-image generation. CoIG utilizes an LLM to decompose a complex prompt into a sequence of simple, step-by-step instructions. The image generation model then executes this plan by progressively generating and editing the image. Each step focuses on a single semantic entity, enabling direct monitoring. We formally assess this property using two novel metrics: CoIG Readability, which evaluates the clarity of each intermediate step via its corresponding output; and Causal Relevance, which quantifies the impact of each procedural step on the final generated image. We further show that our framework mitigates entity collapse by decomposing the complex generation task into simple subproblems, analogous to the procedural reasoning employed by CoT. Our experimental results indicate that CoIG substantially enhances quantitative monitorability while achieving competitive compositional robustness compared to established baseline models. The framework is model-agnostic and can be integrated with any image generation model.

\end{abstract}
    
\section{Introduction}
\label{sec:intro}
  

\begin{figure}[t]
 \centering
     \includegraphics[width=\columnwidth]{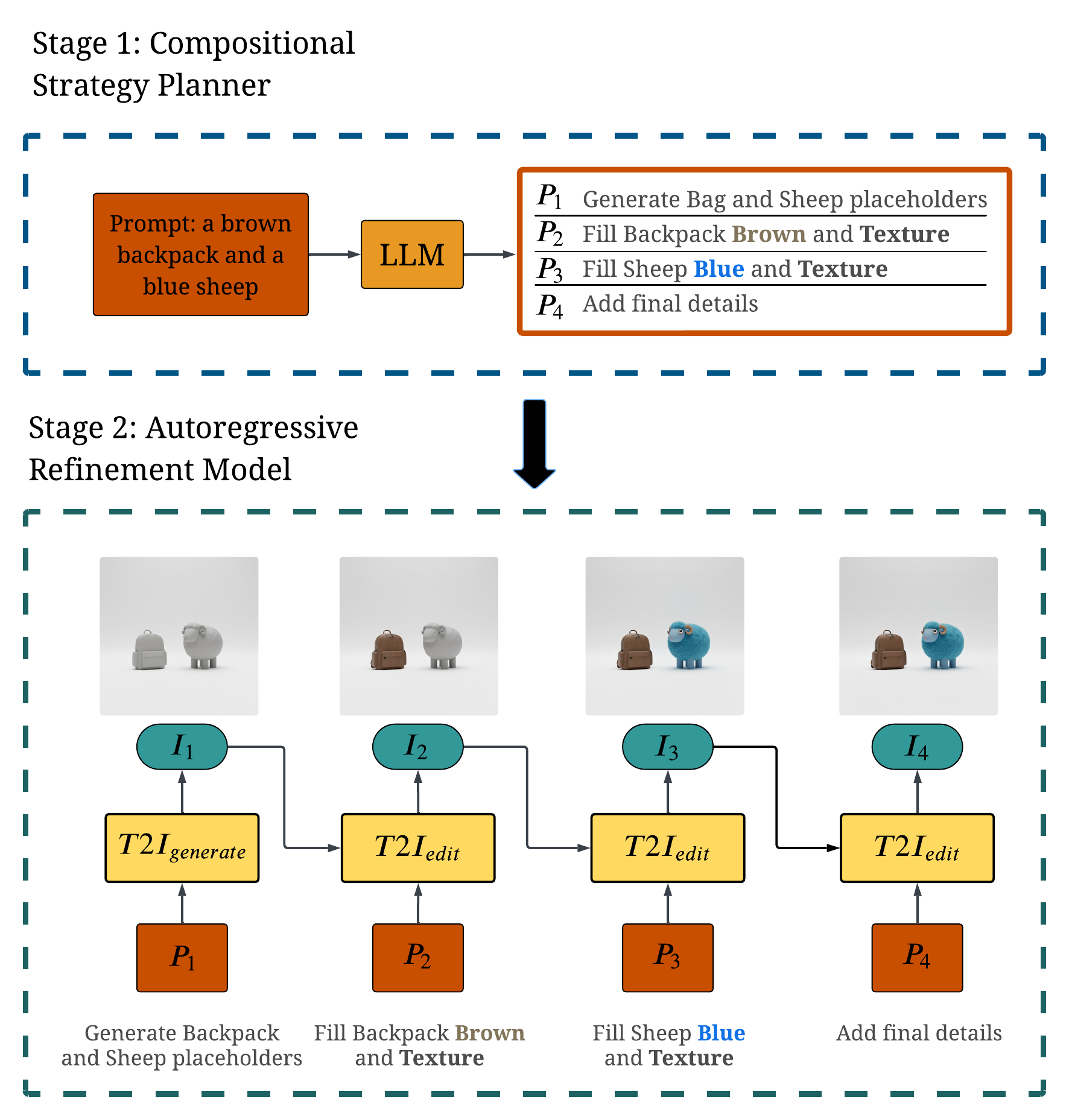} 
 \vspace{-0.6cm}
 \caption{Proposed Chain-of-Image Generation (CoIG) framework. CoIG consists of two principal stages. (1) The Compositional Strategy Planner (CSP), in which an LLM decomposes a single complex prompt into a sequence of simpler sub-prompts $\{P_1, \dots, P_n\}$ that serves as a human-readable and monitorable plan for iterative editing. (2) The Autoregressive Refinement Model (ARM) executes this plan by first generating an initial image using a text-to-image model (T2I) $I_1 = T2I_{generate}(P_1)$ and then iteratively refining the image via $I_t = T2I_{edit}(I_{t-1}, P_t)$, treating the previous image $I_{t-1}$ as an explicit state that conditions subsequent generations.}

 \label{fig:cog_architecture} 
\end{figure}

 
 

 

Recent breakthroughs in generative artificial intelligence (AI), primarily driven by diffusion and autoregressive models, have significantly advanced the field of image synthesis~\cite{sohl2015deep,van2016pixel,van2017neural,parmar2018image,ho2020denoising,song2020score}. The continued scaling of large image-text datasets has dramatically improved the fidelity, versatility, and breadth of generated imagery~\cite{rombach2022high,ramesh2021zero,saharia2022photorealistic,betker2023improving,podell2023sdxl}. Yet, the critical challenge of monitorability remains largely unaddressed despite this rapid progress.

In the broader context of AI, monitorability—defined as the capacity for human (or in some cases, machine) oversight—is fundamental for ensuring accountability, safety, and controllability. This concept is an active area of research within Large Language Model (LLM), where observing Chain-of-Thought (CoT) reasoning pathways is a key area of study~\cite{korbak2025chain,emmons2025chain}.

Extending the principle of monitorability to image generation, however, introduces unique challenges, as dominant synthesis methodologies are fundamentally different from the human approach to creating images. For instance, diffusion models~\cite{ho2020denoising,song2020score} operate by simultaneously refining a complete pixel canvas, while conventional autoregressive models~\cite{li2024autoregressive,lee2022autoregressive} construct an image patch-wise. The human process of image generation, conversely, is inherently both autoregressive and semantic: An artist typically composes a scene by first outlining primary subjects, then gradually adding details to those subjects, and often addressing the background as a final compositional step~\cite{iarussi2013drawing, goel1995sketches}. This semantic progression allows an observer to easily understand the artist's focus at any given stage.

In contrast, the intermediate steps of most generative models are largely non-interpretable to a human observer (or any monitoring system imitating it). The simultaneous refinement of all subjects and background elements in diffusion models obscures clear compositional logic. Similarly, in conventional patch-based autoregressive models, independently determining whether a given patch corresponds to a specific semantic entity or the background is non-trivial. 

To address this limitation, we introduce \textit{Chain-of-Image Generation (CoIG)}, a generation framework that makes image generation monitorable by emulating a human-like process analogous to CoT prompting in reasoning tasks. Building on prior work on LLM-based prompt refinement for compositional synthesis~\cite{yang2024mastering, hao2023optimizing,manas2024improving}, our framework utilizes an LLM to decompose an input prompt into a sequence of sub-prompts. This sequence is designed to mirror a human artist's workflow; for instance, an initial sub-prompt may define the spatial arrangement of all subjects, while subsequent steps progressively elaborate on specific semantic components, addressing one semantic component per step.  The CoIG process operates in an autoregressive manner: An initial image is synthesized from the first sub-prompt, followed by iterative edits based on each succeeding sub-prompt in the sequence. Figure~\ref{fig:cog_architecture} provides an illustrative example of this process.

To quantitatively assess the monitorability of this framework, we introduce two novel metrics for image generation, inspired by similar concepts in LLM evaluation~\cite{korbak2025chain,emmons2025chain}: CoIG Readability, which quantifies the interpretability of intermediate steps; and Causal Relevance, which measures the contribution of each intermediate step to the final generated output. Full metric details are provided in Section~\ref{sec:monitorability}.

Furthermore, as we build CoIG, we identify prevalent failure cases in existing generative models. A notable example occurs when prompts specify multiple, similar entities with different attributes or actions (examples provided later, e.g., Figure~\ref{fig:entity_qual}). In such cases, models often exhibit attribute confusion or merge the entities into a single mode, which we define as ``entity collapse.'' We demonstrate that CoIG mitigates this issue by decomposing the complex task into simpler, sequential steps, similar to how CoT prompting breaks down complex reasoning problems for LLM.

\noindent\textbf{Our main contributions are:}
\begin{itemize}
    \item We introduce and formally address the concept of monitorability within image generation models, highlighting its importance for model reliability, safety, and controllability. To the best of our knowledge, this work provides the first systematic framework for this concept in image generation, extending the critical language step of CoT to the image domain.
    \item We propose CoIG, a novel framework that uses an LLM to decompose a complex prompt into a sequence of tractable, semantic steps.
    \item To quantitatively evaluate monitorability of this and future works on this novel challenge, we introduce two novel metrics for image generation: CoIG Readability, which measures the clarity of intermediate steps to a human observer; and Causal Relevance, which assesses the impact of each step on the final output.
    \item We systematically address ``entity collapse,'' a critical form of attribute binding failure in which attributes of similar entities are merged, and demonstrate its mitigation through our proposed CoIG framework.
\end{itemize}

 
\section{Related Works}
\label{sec:related_works}
\begin{figure*}[t]
 \centering
 
 \includegraphics[width=0.9\textwidth]{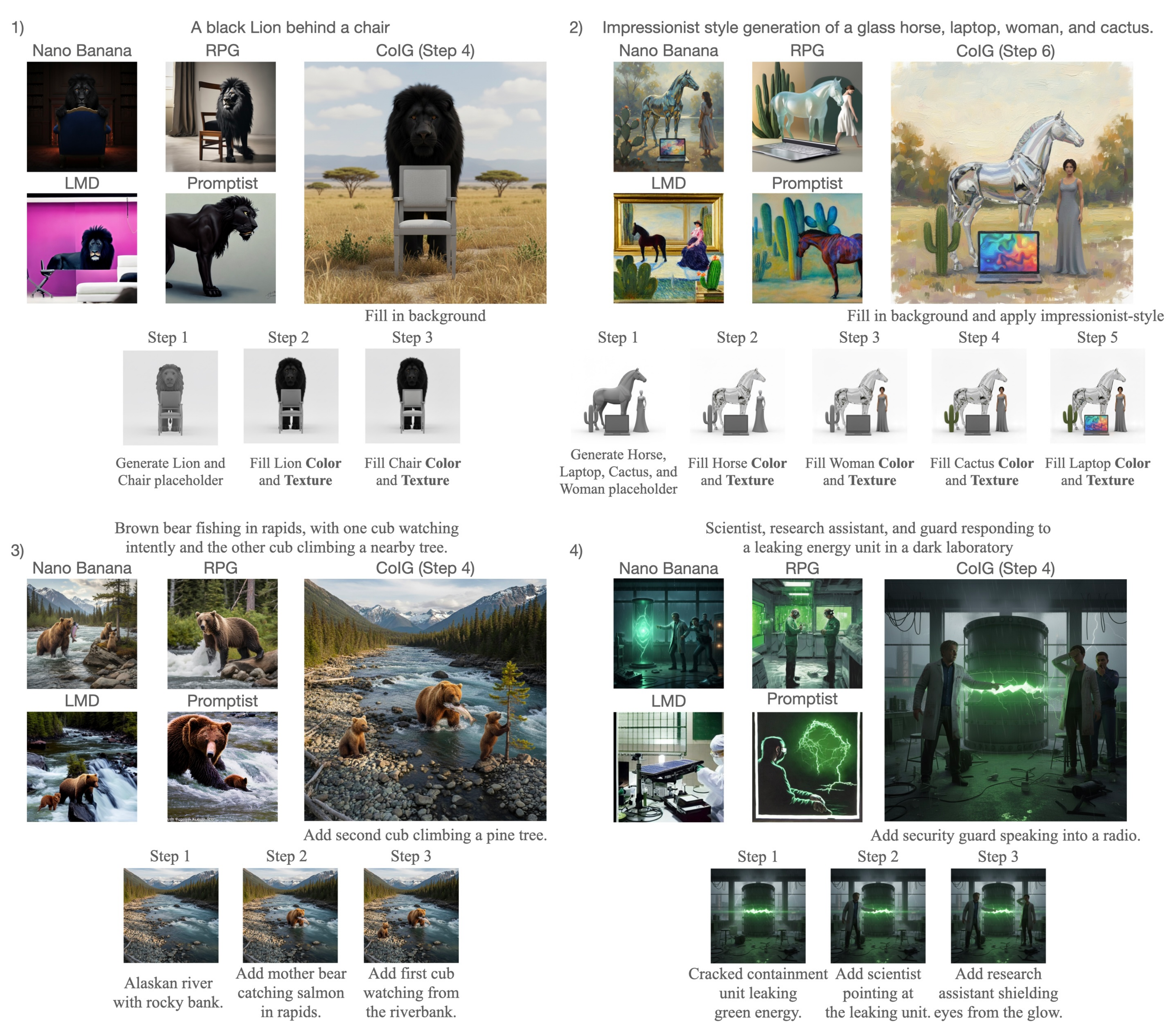}
 
 \caption{Qualitative comparison between our \textbf{CoIG} framework and four baselines—\textbf{Nano Banana}~\cite{fortin2025introducing}, \textbf{RPG}~\cite{yang2024mastering}, \textbf{LMD}~\cite{lian2023llm}, and \textbf{Promptist}~\cite{hao2023optimizing}. Across four complex scenarios that require spatial reasoning and multi-entity coordination (1--4), the baselines frequently struggle with entity collapse and incorrect spatial arrangements. In contrast, CoIG decomposes the prompt into explicit, monitorable steps (shown at the bottom of each panel), ensuring precise object placement and correct attribute binding. For additional examples, please see the supplementary material in Appendix~\ref{app:add_exp}.}
 \vspace{-0.28cm}
 \label{fig:CoIG_Read_qual}
\end{figure*}

A key advancement for LLMs is CoT prompting, a technique that enables models to reason in a step-by-step manner similar to humans before providing an output~\cite{wei2022chain, kojima2022large,lu2022dynamic,yao2023tree,zhao2023verify}. This technique has proven effective for a range of complex tasks, including arithmetic problem-solving, symbolic reasoning, and medical question-answering~\cite{wang2022self, lewkowycz2022solving, yao2023tree, singhal2023large, singhal2025toward}. The emergence of these explicit rationales prompted critical research into their faithfulness, whether the reasoning is causally linked to the model's output or is merely a post hoc justification~\cite{jacovi2020towards,turpin2023language,arcuschin2503chain,chen2025reasoning}. Moving beyond post hoc analysis, the concept of monitorability also emerged, referring to the capacity for a human to observe and intervene in the generative process—a crucial property for ensuring reliability, safety, and control~\cite{korbak2025chain,emmons2025chain}. Despite these advancements in the language domain, these principles have not yet been translated to image generation. This is primarily because widely used models, such as diffusion and autoregressive models, operate in a non-semantic manner that is inherently difficult for a human to monitor. This architectural limitation motivates the need for a new framework capable of producing a human-readable generative process for images.

A central challenge in text-to-image synthesis is compositionality, meaning the ability to faithfully render multiple entities with specified attributes and relationships~\cite{liu2022compositional,huang2023composer,zheng2023layoutdiffusion,li2023gligen}. The growing focus on this issue has led to the development of a variety of dedicated benchmarks designed to systematically evaluate this skill, with prominent examples including T2I-CompBench and others~\cite{huang2023t2i,ghosh2023geneval,li2024genai,hu2024ella,wu2024conceptmix,wei2025tiif,li2025easier}. These benchmarks provide a diverse suite of prompts that test challenging abilities, including complex attribute binding and spatial relationships. In addition to these known challenges, we investigate a subtle but critical failure mode that we refer to as ``entity collapse.'' This phenomenon describes a specific form of attribute binding failure in which models incorrectly merge features between similar entities, an issue that is related to but distinct from the broader compositional problems tested.

In response, a significant body of research has aimed to address these issues. One prominent approach leverages Large Language Models (LLMs) to rephrase or decompose complex prompts into more structured inputs that provide more granular guidance for image generation~\cite{yang2024mastering, hao2023optimizing, lian2023llm}. Concurrently, another line of work provides granular control over the final image by accepting spatial guidance inputs. Frameworks like ControlNet~\cite{zhang2023adding} and T2I-Adapter~\cite{mou2024t2i} use inputs such as human poses, depth maps, or Canny edges to guide the spatial layout and structure of the final image. However, while these methods significantly improve control over the final output and thus enhance reliability, they do not make the generative process itself transparent. The user can guide the outcome but cannot inherently observe the model's step-by-step construction of the image, leaving a critical gap in providing a generation process that is both controllable and directly monitorable.

\section{Chain-of-Image Generation Framework}
\label{sec:CoIG}

The CoIG framework is designed to enhance the monitorability of the image generation pipeline. Our approach consists of two core components: A Compositional Strategy Planner (CSP), leveraging an LLM to decompose the prompts; and an Autoregressive Refinement Model (ARM), employing a text-to-image model (T2I) for iterative image synthesis. This two-stage process transforms a single, complex prompt into a monitorable, step-by-step generation process that mimics the human artistic process. The framework pipeline is depicted in Figure~\ref{fig:cog_architecture}.

\subsection{Compositional Strategy Planner}
\label{subsec:CSP}



The CSP transforms conventional one-step generation into a transparent, sequential process. We implement CSP using a pre-trained LLM, prompted with specific decomposition rules (described in Appendix~\ref{app:exp_setup}), to parse the original prompt into a logically ordered sequence of $n$ simplified sub-prompts ${P_1, P_2, \dots, P_n}$, each focusing on a single semantic component (e.g., the color or shape of a specific object). This decomposition serves two primary functions:

\noindent\textbf{Ensuring monitorability.} The sequence of sub-prompts serves as a human-readable  ``plan of action.'' Before image generation, one can inspect this plan and supervise the model's intended trajectory. Such transparency underpins monitorability by enabling verification and intervention, thereby enhancing model reliability and safety.

\noindent\textbf{Enforcing compositionality.} By separating a complex scene into discrete tasks, we prevent the model from ``blending'' or ``collapsing'' attributes between similar entities, as handling each attribute in a separate step. This is demonstrated in the Figure~\ref{fig:CoIG_Read_qual}.

\subsection{Autoregressive Refinement Model}
\label{subsec:ARM}

The ARM uses the generated sequence of sub-prompts $\{P_1, P_2, \dots, P_n\}$ to progressively guide the model toward the intended final image. The initial image $I_1$ is directly generated from $P_1$, written as
\begin{equation}
    I_1 = T2I_{\text{generate}}(P_1),
\end{equation}
where $T2I_{\text{generate}}(\cdot)$ denotes the T2I model's generative operation. This initial image establishes the basic composition, such as a placeholder for each subject or the background of the scene. 

In each subsequent step, the T2I model refines previous output $I_{t-1}$ according to the current sub-prompt $P_t$,
\begin{equation}
    I_t = T2I_{\text{edit}}(I_{t-1}, P_t),
\end{equation}
where $T2I_{\text{edit}}(\cdot)$ denotes the T2I model's editing operation. Each intermediate image $I_{t-1}$ thus functions as an explicit state, carrying the cumulative visual context up to step $t-1$. This mechanism is analogous to state propagation in conventional autoregressive models, but with each state fully observable and monitorable. When necessary, we can edit $P_t$ to guardrail the generative process within our intended trajectory.

\section{Monitorability in Image Generation}
\label{sec:monitorability}

In generative AI, particularly concerning LLMs, monitorability is predicated on two fundamental principles~\cite{korbak2025chain,emmons2025chain}: (1) \textbf{Readability}, the capacity of a model to produce interpretable reasoning; and (2) \textbf{Causal Relevance}, the existence of a causal linkage between this reasoning and the final output. If the reasoning lacks this causal relationship, monitoring it is meaningless, as intervening in or altering intermediate steps does not guarantee changing the final outcome~\cite{adebayo2018sanity,carloni2025role}. While a monitorable LLM reasoning mimics human cognition, a monitorable image-generation process should likewise approximate human image creation. Accordingly, we provide formal definitions of \textbf{Readability} and \textbf{Causal Relevance} in CoIG,



\begin{defbox}{Readability of CoIG}\label{def:coig-readability}
\textbf{Readability of CoIG} requires that each intermediate generative step be \textbf{visually self-explanatory}. By requiring that each update introduce a single distinct visual concept—whether a specific object, a coherent group of entities, or a targeted attribute—we ensure that users can discern the purpose of each step simply by observing the image. This restriction minimizes ambiguity, allowing errors to be immediately localized to specific instructions rather than obscured within complex, multifaceted updates. Figure~\ref{fig:readability_qual2} illustrates this principle.
\end{defbox}

\begin{figure}[t]
 \centering
 
 \includegraphics[width=0.85\columnwidth]{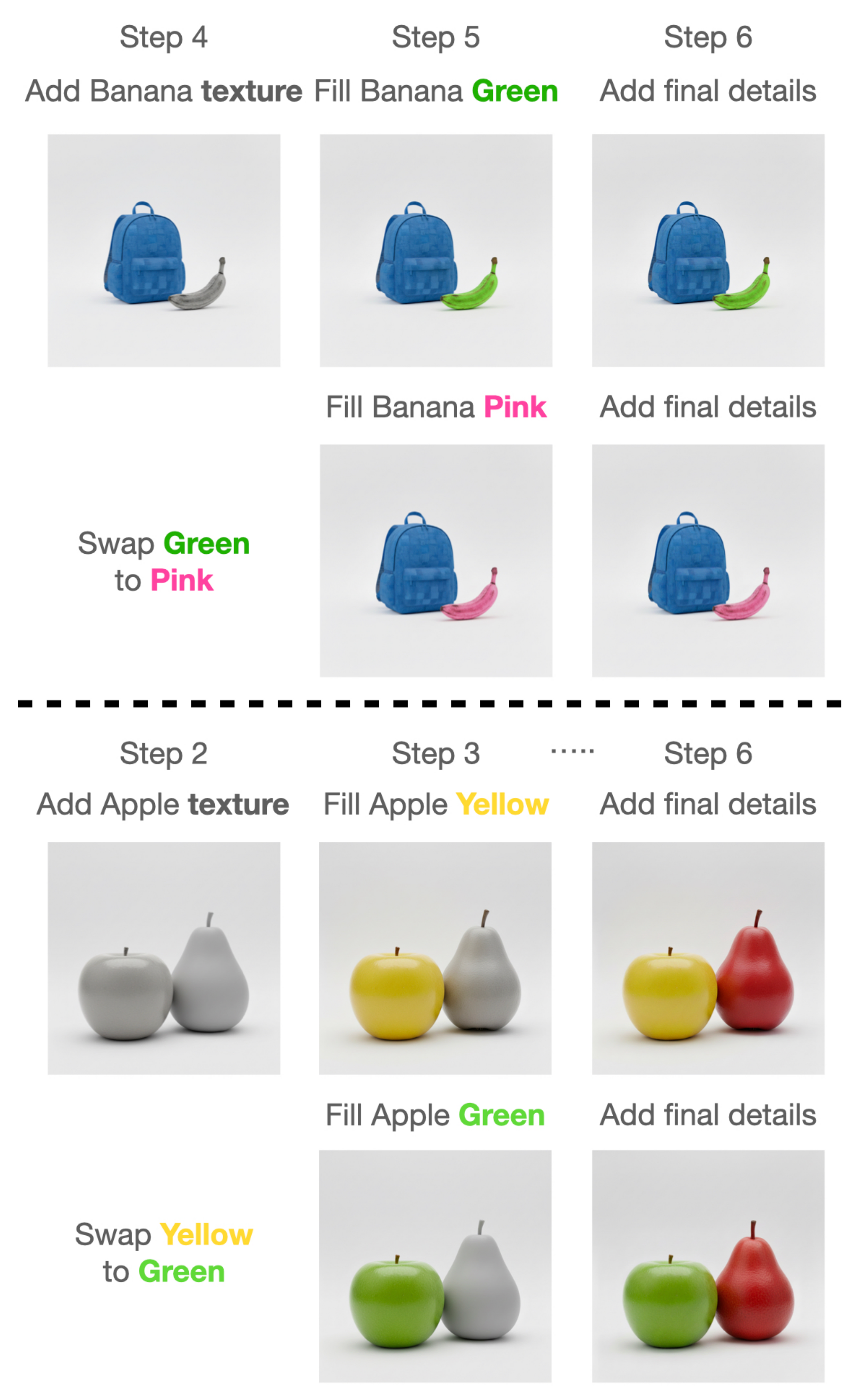}
 \vspace{-0.23cm}
 \caption{Qualitative demonstration of Causal Relevance. Rows 1 and 3 show the final images from two original CoIG sequences. Rows 2 and 4 show final images resulting from single perturbations applied during intermediate steps, which change the color of the apple (row 2) and the banana (row 4). The persistence of these targeted changes confirms the causal link between intermediate steps and the final output.}
 \label{fig:causal_qualitative}
 \vspace{-0.3cm}
\end{figure}

\begin{defbox}{Causal Relevance}\label{def:coig-causal-relevance}
\textbf{Causal Relevance} requires each intermediate generative step to contribute demonstrably and persistently to the final output. Analogous to an artist’s brushstroke, once a subject is introduced or a detail refined, its effect should persist through subsequent stages. If an action’s effects disappear or are overwritten by the final step, that step lacks causal relevance and becomes meaningless to monitor. Figure~\ref{fig:causal_qualitative} illustrates this principle.
\end{defbox}

Our proposed framework implements these two properties through \textit{compositional lock}, which ensures that content generated in prior steps is locked and remains unaltered unless explicitly targeted by the current instruction~\cite{hertz2022prompt,simsar2025lime}. Readability is secured by altering only the prompted subject while keeping the rest fixed, making the visual change clearly tied to the task. Causal relevance is ensured by preserving all prior edits, which yields a visual analog of faithfulness—whether the model's explicit reasoning causally produces the output rather than a post hoc justification after the output is generated~\cite{turpin2023language,chen2025reasoning}. This makes the process meaningful for human monitoring.

\begin{table*}[!ht]
\centering
\footnotesize
\setlength{\tabcolsep}{4pt}

\subcaptionbox{Comparison on the GenEval benchmark.\label{tab:geneval_results}}[0.325\textwidth]{%
  \begin{tabular}{lcc}
    \toprule
    \textbf{Metric} & \textbf{Baseline~\cite{huang2023t2i}} & \textbf{CoIG (ours)} \\
    \midrule
    Single Obj.   & 83.75 & 98.75 \\
    Two Obj       & 90.91 & 98.99 \\
    Counting      & 92.50 & 86.25 \\
    Colors        & 88.30 & 92.55 \\
    Position      & 92.50 & 99.00 \\
    Color Attri.  & 91.50 & 93.50 \\
    Overall       & 89.91 & 94.84 \\
    \bottomrule
  \end{tabular}%
}
\subcaptionbox{Comparison on the T2I-CompBench benchmark.\label{tab:compbench_results}}[0.325\textwidth]{%
  \begin{tabular}{lcc}
    \toprule
    \textbf{Metric} & \textbf{Baseline~\cite{huang2023t2i}} & \textbf{CoIG (ours)} \\
    \midrule
    Color        & 90.43 & 93.31 \\
    Shape        & 82.92 & 84.02 \\
    Texture      & 83.57 & 83.85 \\
    Spatial      & 91.10 & 94.83 \\
    Non-Spatial  & 86.60 & 84.77 \\
    Complex      & 70.63 & 69.69 \\
    Overall      & 84.20 & 85.08 \\
    \bottomrule
  \end{tabular}%
}
\subcaptionbox{Comparison on the ConceptMix benchmark.\label{tab:conceptmix_results}}[0.325\textwidth]{%
  \begin{tabular}{lcc}
    \toprule
    \textbf{Metric} & \textbf{Baseline~\cite{huang2023t2i}} & \textbf{CoIG (ours)} \\
    \midrule
    $k=1$      & 91.76 & 91.04 \\
    $k=2$      & 89.26 & 86.91 \\
    $k=3$      & 88.02 & 84.19 \\
    $k=4$      & 84.70 & 78.58 \\
    $k=5$      & 83.42 & 81.17 \\
    $k=6$      & 83.64 & 79.55 \\
    $k=7$      & 84.11 & 78.49 \\
    Overall    & 86.43 & 82.86 \\
    \bottomrule
  \end{tabular}%
}
\vspace{-0.25cm}
\caption{Comparison of our proposed CoIG and the baseline across three benchmarks. While the main goal of the proposed framework is measured qualitatively, these quantitative results show competing performance with state-of-the-art, giving confidence to the approach.}
\label{tab:all_results}
\vspace{-0.25cm}
\end{table*}

We quantitatively evaluate the Readability and Causal Relevance of CoIG via the following methods:

\begin{enumerate}
    \item \textbf{Evaluation of CoIG Readability.}
    We use a Multimodal Large Language Model (MLLM) as an automated proxy for human judgment of Readability (described in detail in Appendix~\ref{app:exp_setup}).
    For any intermediate image $I_t$ generated after applying sub-prompt $P_t$ targeting a semantic entity $e_t$, the MLLM verifies the presence and attributes of $e_t$ in $I_t$.
    High fidelity between $P_t$ and the MLLM's verification indicates high readability.

    \item \textbf{Evaluation of CoIG Causal Relevance.}
    We evaluate Causal Relevance via controlled perturbation to sub-prompts. Specifically, we alter $P_t$ (e.g., ``red bowl'' to ``blue bowl'') and evaluate (i) whether the change appears in the corresponding intermediate image $I_t$, and (ii) whether the change persists through subsequent images, especially the final output $I_n$. A modification that is both instantiated at $t$ and maintained to $n$ indicates a direct, durable causal effect on the final image.

\end{enumerate}

\section{Compositional Robustness and Entity Collapse}
\label{sec:entity_collapse}

In this section, we formally define ``entity collapse,'' a critical failure mode in compositional text-to-image generation that represents a particularly complex form of the attribute binding problem. We analyze its causes and introduce a new benchmark, the Entity Collapse (EC) Benchmark, designed for the systematic evaluation of this phenomenon.

\subsection{Defining and Analyzing Entity Collapse} 


\begin{defbox}{Entity collapse}\label{def:entity-collapse}
\textbf{Entity collapse} occurs when a prompt specifies \(n\) semantically similar entities with distinct attributes \(\{A_i\}_{i \in \{1, \dots, n\}}\), but the generated image (i) depicts fewer than \(n\) instances (merge), (ii) misassigns attributes (swap/leak), or (iii) applies one entity’s attributes to all (homogenization). Visual examples are illustrated in Figure~\ref{fig:entity_qual}.
\end{defbox}

This failure is prevalent in complex prompts involving entities from an identical semantic class (e.g., ``four airmen'') with distinctive attributes (e.g., one ``smiling,'' one ``wearing glasses''). We attribute this failure to the excessive compositional burden imposed by a single-pass generation paradigm~\cite{zarei2024understanding,campbell2024understanding}. Conventional diffusion or autoregressive models are constrained to generate the entire image from a complex prompt within a single inference. This constraint imposes a processing load that overloads the model's compositional capacity, making it difficult to differentiate between distinct, semantically related entities and thereby resulting in the observed collapse of attributes. Examples of collapse are illustrated in Figure~\ref{fig:entity_qual}.

\subsection{Mitigation of Entity Collapse via Chain-of-Image Generation}

Our proposed CoIG framework mitigates this by reframing the complex task as a sequence of tractable sub-tasks, a strategy analogous to CoT reasoning. CoIG generates a sequential plan: First, it creates placeholders for each entity while considering their interactions. Subsequent steps progressively fill in details for specific attributes and interactions. The final step then fills in a contextually coherent background. This sequential execution, combined with the \emph{compositional lock} (see Section~\ref{sec:monitorability}), intentionally constrains the model's computational focus. At each step, the model processes only a single entity or interaction. This drastically reduces the compositional load, preventing the attribute-binding failures characteristic of entity collapse.

\subsection{The Entity Collapse (EC) Benchmark}
\label{sec:ec_benchmark}

To systematically evaluate robustness to entity collapse, we developed the \textit{EC Benchmark} (to be released with this paper), a new 300-prompt benchmark specifically designed to induce this failure mode. This benchmark is employed in Section~\ref{sec:experiments} to conduct a quantitative comparison. Full details on the procedural prompt generation and vocabularies are in Appendix~\ref{app:ec_benchmark_detail}.

\section{Experiments}
\label{sec:experiments}

In this section, we present a comprehensive evaluation of our proposed CoIG framework.
Our primary baseline for comparative evaluations is Google Nano Banana~\cite{fortin2025introducing}, used in a single pass in contrast to the multi-step CoIG methodology.
For the experiments presented in this section, our proposed CoIG framework was implemented using Gemini 2.5 Flash~\cite{comanici2025gemini} as the Compositional Strategy Planner and Google Nano Banana~\cite{fortin2025introducing} as the Autoregressive Refinement Model.
Full details on our experimental setup, including evaluation protocols and benchmark descriptions, are available in Appendix~\ref{app:exp_setup}.

\subsection{Evaluation on Compositionality}
\label{subsec:exp_compositionality}

Figure~\ref{fig:CoIG_Read_qual} presents a qualitative comparison against four baselines: \textbf{Nano Banana}~\cite{fortin2025introducing}, \textbf{RPG}~\cite{yang2024mastering}, \textbf{LMD}~\cite{lian2023llm}, and \textbf{Promptist}~\cite{hao2023optimizing}. The results highlight that the baselines struggle with \textit{entity collapse} in complex prompts—for example, failing to distinguish the specific actions of two bear cubs (Panel 3) or blending the distinct roles of laboratory personnel (Panel 4). In contrast, CoIG’s stepwise decomposition ensures high-fidelity adherence to spatial and semantic constraints, while preserving both monitorability and overall generative performance. To demonstrate the general applicability of our framework, we provide an additional qualitative ablation study across several distinct LLM and image generation backbones in Appendix~\ref{app:generalizability}.

We evaluate the proposed CoIG framework on the standard GenEval, T2I-CompBench, and ConceptMix benchmarks~\citep{huang2023t2i,ghosh2023geneval,wu2024conceptmix} to establish its effectiveness as a general-purpose generation method. Full details on those datasets are in Appendix~\ref{app:exp_setup}. For our evaluation protocol, we use an MLLM evaluator adapted from~\cite{li2025easier}. This evaluator is prompted to verify if the prompt's compositional instructions are accurately rendered in the generated image, providing binary (yes/no) answers. 




The quantitative results for all three benchmarks are presented in Table~\ref{tab:all_results} (while the key contribution of our work is evaluated in a qualitative fashion, these quantitative evaluation further supports the proposed framework). We first present the GenEval results in Table~\ref{tab:geneval_results}. CoIG achieves a better overall score (94.84 vs. 89.91) and outperforms the baseline in the majority of categories, including Single Object, Two Objects, Position, and Color Attribute. Notably, it underperforms the baseline in the ``Counting'' category.

The evaluation on T2I-CompBench, detailed in Table~\ref{tab:compbench_results}, shows a more nuanced outcome. While CoIG demonstrates improved performance in the Color and Spatial categories, it lags behind the baseline in Non-Spatial and Complex compositions. The results for Shape and Texture are comparable, leading to a slightly higher overall score (85.08 vs. 84.20), while the qualitative examples in Figure~\ref{fig:readability_qual2} further indicate that CoIG tends to produce more faithful compositions than the baseline.

\begin{figure}[t]
 \centering
 
 \includegraphics[width=\columnwidth]{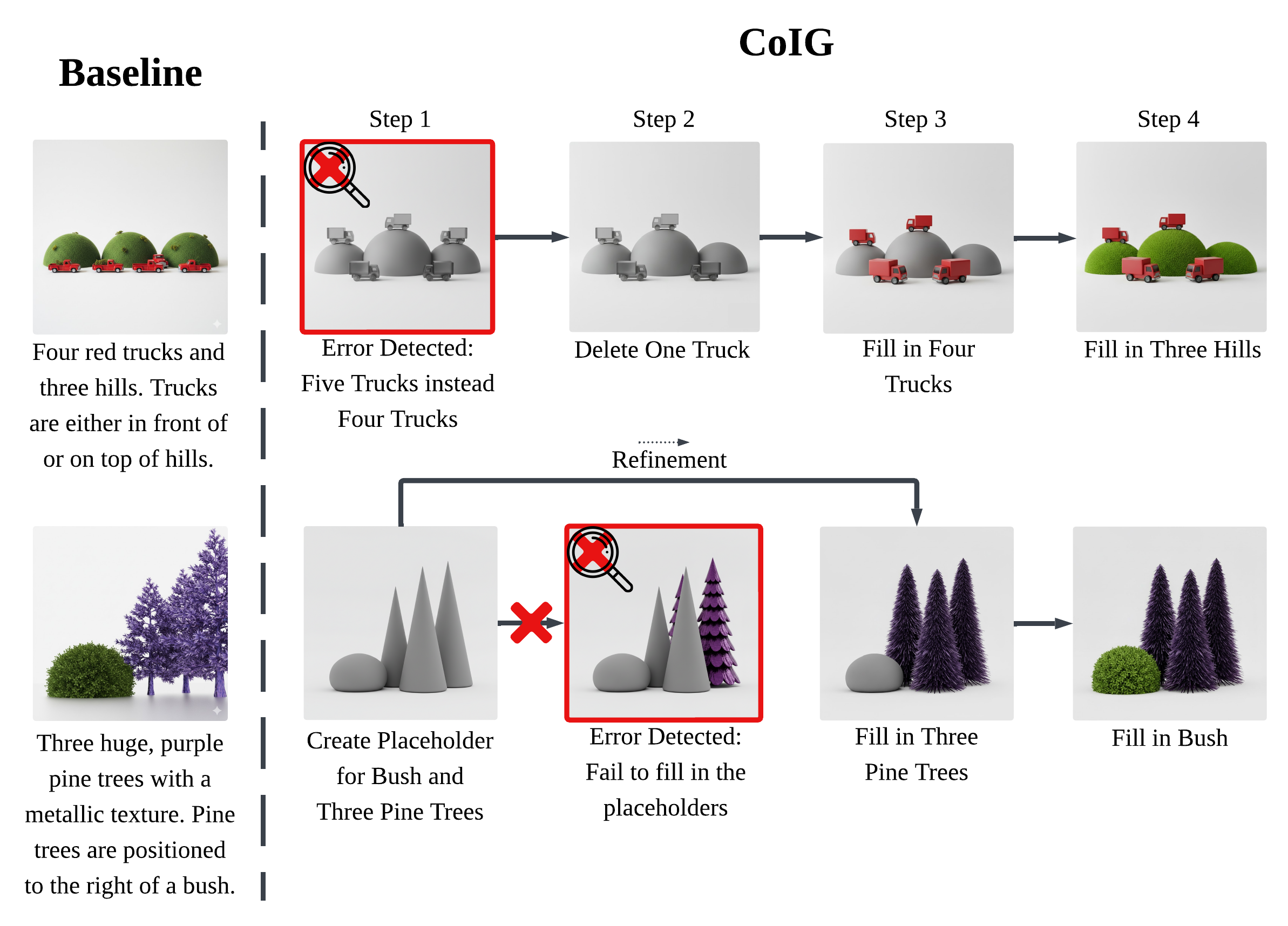}
 \vspace{-0.5cm}
 \caption{\textbf{Leveraging monitorability for error correction.} CoIG’s stepwise structure exposes intermediate failures, allowing a monitor to intervene. Top: A counting error (five trucks instead of four) is detected in the layout phase and corrected via a targeted ``Delete One Truck'' instruction. Bottom: A failure to faithfully fill the placeholder for the pine trees is identified, triggering a refinement step that correctly renders the ``purple pine trees.''}
 \label{fig:monitor_intervention}
 \vspace{-0.2cm}
\end{figure}

On ConceptMix (Table~\ref{tab:conceptmix_results}), CoIG underperforms the baseline across the reported $k$ values. However, we leverage CoIG’s monitorable, stepwise structure to diagnose the source of this gap. We observe that the performance drop stems from the zero-shot nature of our layout generation. While off-the-shelf models generalize well enough to create placeholders for spatially distinct objects in standard benchmarks, they lack the explicit supervision required to handle dense compositions and overlapping attributes specific to ConceptMix. This limitation manifests as two failure modes when scene complexity exceeds the model's zero-shot capabilities: (i) the base generator produces ambiguous placeholders for crowded entities, and (ii) the ARM struggles to ``fill in'' these regions without leaking. Crucially, the framework's transparency allows us to catch these model-specific failures. Because each error is localized, a human or MLLM monitor can intervene to correct the instruction, as illustrated in Figure~\ref{fig:monitor_intervention}.

\begin{table}[h]
\centering
\begin{tabular}{lrrr}
\toprule
  & \textbf{Color} & \textbf{Shape} & \textbf{Texture} \\
\midrule
\textbf{Before} & 13.12 & 59.70 & 8.65 \\
\textbf{After} & 88.71 & 81.13 & 74.68 \\
\bottomrule
\end{tabular}%
\caption{Quantitative evaluation of CoIG readability. We compare MLLM evaluator scores across three semantic categories (Color, Shape, and Texture) ``Before'' and ``After'' the corresponding detailing step. The results show a consistent performance increase across all attributes.}
\label{tab:CoIG_Read}
\end{table}

\subsection{Evaluation of Monitorability}
\label{subsec:exp_monitorability}

\subsubsection{Readability of the Chain-of-Image Generation}


We evaluate readability on T2I-CompBench as detailed in Section~\ref{sec:monitorability}. The quantitative results in Table~\ref{tab:CoIG_Read} confirm the high readability of our steps. For Color and Texture, we observe near-zero initial detection followed by a sharp post-step surge, indicating that the generated details are distinct and immediately attributable to the corresponding steps. In contrast, Shape begins with a higher baseline ($59.70$), since the placeholder already establishes the object's structural form; however, the subsequent increase to $81.13$ remains substantial, confirming effective refinement.


\begin{figure}[h]
 \centering
 
 \includegraphics[width=\columnwidth]{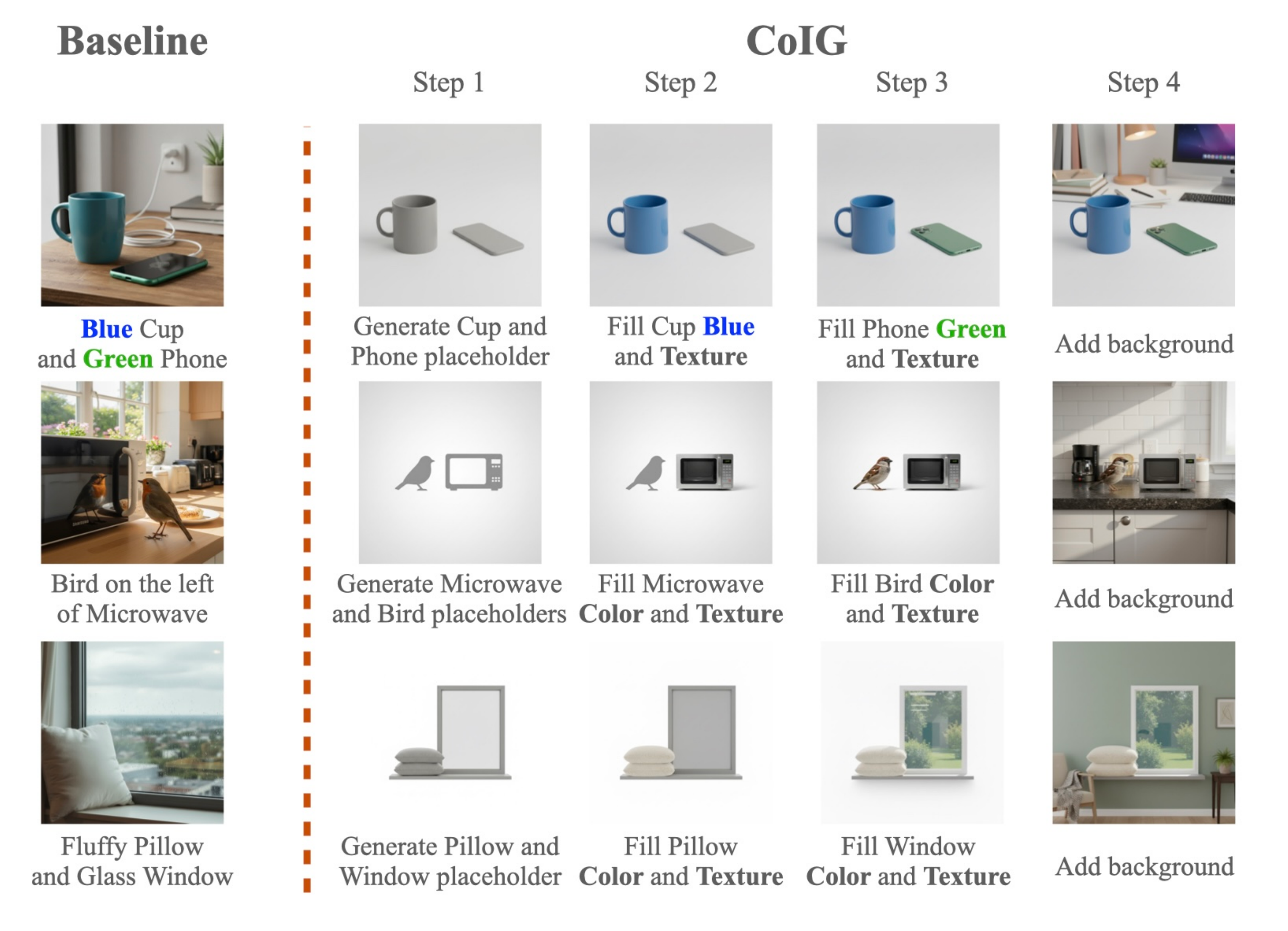}
 
\vspace{-0.2cm}
 \caption{Visualizing the monitorable advantage of CoIG. In contrast to the black box Baseline, which suffers from attribute leakage (e.g., bleeding ``green'' onto the cup in row 1) or ignores texture constraints (e.g., missing the ``fluffy'' texture in row 3), CoIG's stepwise process ensures precise attribute binding. By isolating semantic components into readable steps, our framework provides direct control correcting these failures. More examples demonstrating readability are shown in Appendix~\ref{app:readbility_more}.}
 \label{fig:readability_qual2}
 
\end{figure}



Figure~\ref{fig:readability_qual2} provides a qualitative comparison, demonstrating the readable and monitorable nature of our proposed CoIG framework, especially compared to the baseline. This monitorable chain stands in sharp contrast to the black box nature of the baseline, and our framework can also flexibly adapt the chain length based on task complexity.

\subsubsection{Causal Relevance}



For evaluating the causal relevance of CoIG, we evaluated on the Color datasets from T2I-CompBench using the method described in Section~\ref{sec:monitorability}. Notably, we avoided perturbations that involved changing an object's color to gray. This is because our image generation model uses gray to render placeholders, and the MLLM evaluator might consequently identify this gray placeholder as the intended object, skewing the evaluation.

In the original, unperturbed sequence, the MLLM score for the new attribute was, as expected, near zero (1.35). When we introduced a perturbation at an intermediate step in the sequence, the score for the new attribute increased substantially at that step to 85.32 and, crucially, remained high in the final output, with a score of 89.52. This pattern indicates that modifications at intermediate steps in our proposed CoIG framework have a direct and meaningful causal impact on the final image.

This causal link is further confirmed in our qualitative results shown in Figure~\ref{fig:causal_qualitative}. The figure visually illustrates the impact of single-prompt perturbations. These examples visually confirm that modifications made during intermediate steps are carried through to the final output, demonstrating the high causal relevance of the CoIG process and supporting the efficacy of monitoring and intervening. More examples are shown in Appendix~\ref{app:qual_causal}.

\begin{table}[t]
\centering
\resizebox{\columnwidth}{!}{%
\begin{tabular}{lrr}
\toprule
\textbf{Metric} & \textbf{Baseline} & \textbf{CoIG (ours)} \\
\midrule
Entity Count (out of 1) & 0.724 (72.4\%) & 0.877 (87.7\%) \\
Attribute Binding (out of 4) & 3.312 (82.8\%) & 3.279 (82.0\%) \\
Interaction (out of 2) & 0.980 (49.0\%) & 1.292 (64.6\%) \\
\midrule
\textbf{Total Score (out of 7)} & \textbf{5.017 (71.7\%)} & \textbf{5.449 (77.8\%)} \\
\bottomrule
\end{tabular}%
} 
\vspace{-0.2cm}
\caption{Quantitative comparison on the Entity Collapse (EC) Benchmark. Our model shows significant improvements in entity count and interaction scores, leading to a higher overall score.}
\label{tab:ec_benchmark_results}
\vspace{-0.3cm}
\end{table}

\subsection{Evaluation on the Entity Collapse (EC) Benchmark}
\label{subsec:exp_entity_collapse}

\begin{figure}[!ht]
 \centering
 
 \includegraphics[width=\columnwidth]{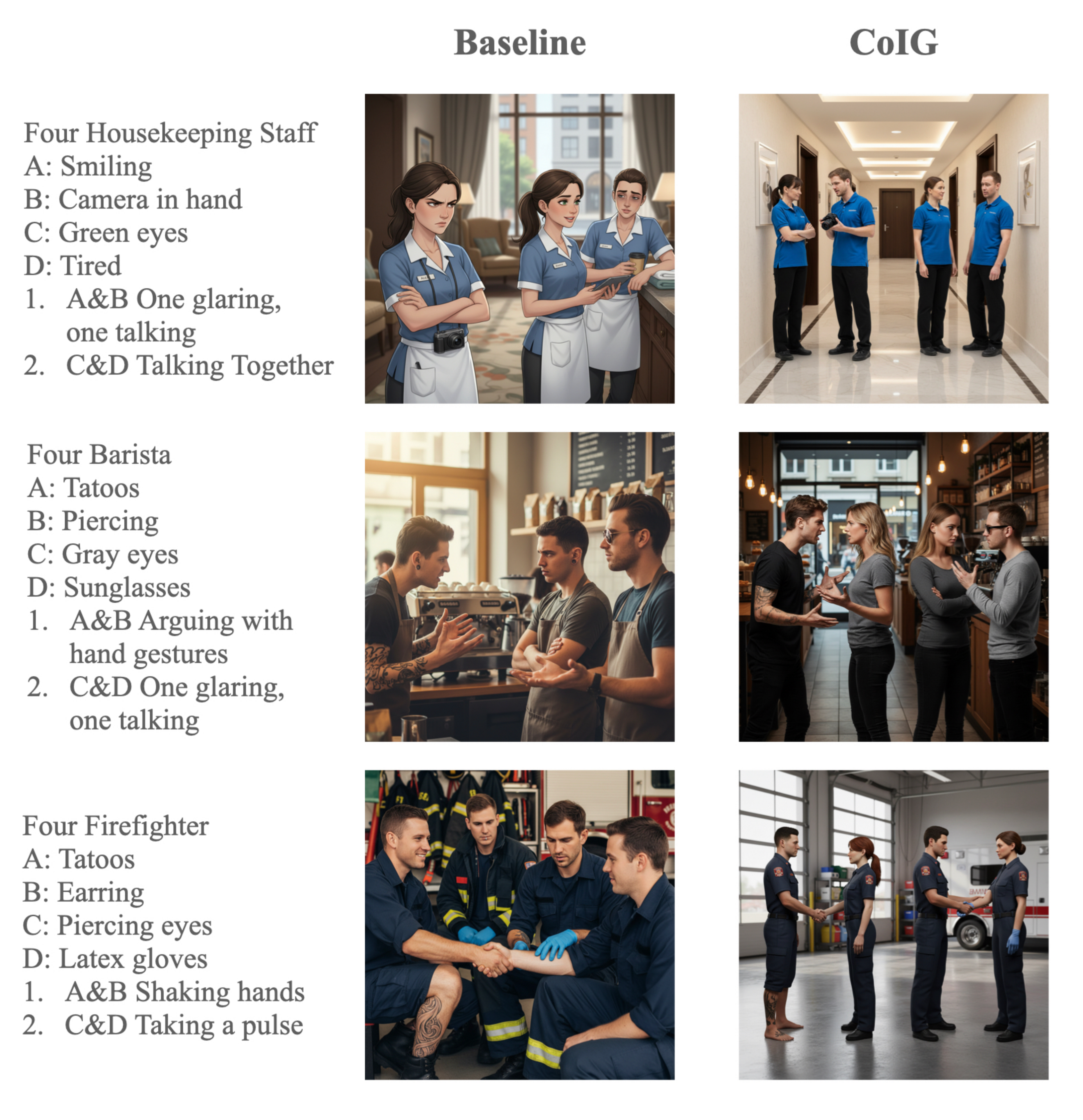}
 \vspace{-0.6cm}
 \caption{A qualitative comparison illustrating how CoIG mitigates \textbf{entity collapse}. The baseline model (center) exhibits the ``merge'' failure from Definition~\ref{def:entity-collapse} in Rows 1 and 2, generating only \textbf{three} entities instead of the prompted \textbf{four}. In Row 3, while the baseline correctly generates four entities, it exhibits a critical \textbf{interaction assignment failure}, it fails to assign exactly one interaction to each entity, instead assigning two interactions (``Shaking hands'' and ``Taking a pulse'') to a single entity. In contrast, our proposed CoIG framework (right) successfully decomposes the task in all rows, generating all four distinct entities and faithfully rendering their complex, paired interactions. More examples are shown in Appendix~\ref{app:entity_qual2}.}
 
 \label{fig:entity_qual}
\end{figure}

We test whether CoIG’s sequential generation process can mitigate the ``entity collapse'' failure mode by evaluating it against the Gemini baseline on a 300-prompt EC benchmark, using Entity Count, Attribute Binding, and Interaction correctness as metrics. CoIG achieves a higher total score (5.449 vs. 5.017), mainly due to a 21.1\% gain in Entity Count and a 31.8\% gain in Interaction scores, indicating that it is more robust to entity collapse and better at rendering relationships between entities. These trends are also visible in the qualitative examples: The baseline often collapses multiple described entities into a single object or misses the specified interaction, while CoIG produces distinct entities with the intended interactions, as illustrated in Figure~\ref{fig:entity_qual}.

The baseline is slightly stronger on Attribute Binding, which we hypothesize is because it can sometimes attach several attributes to the few entities it does generate, even when it fails to produce the right number of entities or interactions. In contrast, CoIG’s sequential, placeholder-based generation explicitly allocates one slot per entity, helping maintain a one-to-one mapping between entities and attributes. Although the iterative editing step can still make mistakes and CoIG is not universally strong at counting or filling in placeholders, as discussed in Section~\ref{subsec:exp_compositionality}, it more reliably avoids the severe failure of distinct entities merging into one, making it substantially more robust to entity collapse than the baseline.


\section{Conclusion}
\label{sec:conclusion}

This research addresses the critical challenge of monitorability in image generation—a property fundamental to model reliability and safety, yet one that has been largely unaddressed by ``black box'' generative architectures. We proposed the CoIG framework, a novel approach that reframes image generation as a sequential, human-like semantic process. By leveraging an LLM to decompose complex textual prompts into discrete, incremental sub-tasks, CoIG creates an inherently transparent and interpretable generation workflow.

Our empirical evaluation yields three primary findings. First, CoIG achieves competitive performance on standard compositionality benchmarks and establishes monitorability without compromising generative quality. Second, we quantitatively validate our framework's monitorability using two novel metrics, CoIG Readability and Causal Relevance, confirming that our intermediate steps are both interpretable and have a faithful impact on the final image. Third, CoIG effectively mitigates the ``entity collapse'' failure mode, significantly outperforming the baseline on our newly introduced EC Benchmark by accurately rendering distinct entities and their specified interactions.

By shifting the paradigm from single-pass generation to a monitorable, sequential semantic process, we argue that monitorability and compositional robustness are deeply intertwined. Our work lays the foundation for a new class of more reliable and trustworthy generative models.

\section*{Acknowledgments}
Work partially supported by ONR, NSF, and Simons Foundation.

\clearpage
{
    \small
    \bibliographystyle{ieeenat_fullname}
    \bibliography{main}
}

\clearpage
\setcounter{page}{1}
\maketitlesupplementary


\section{Additional Qualitative Results}
\label{app:add_exp}

\begin{figure*}[h]
 \centering
 
 \includegraphics[width=\textwidth]{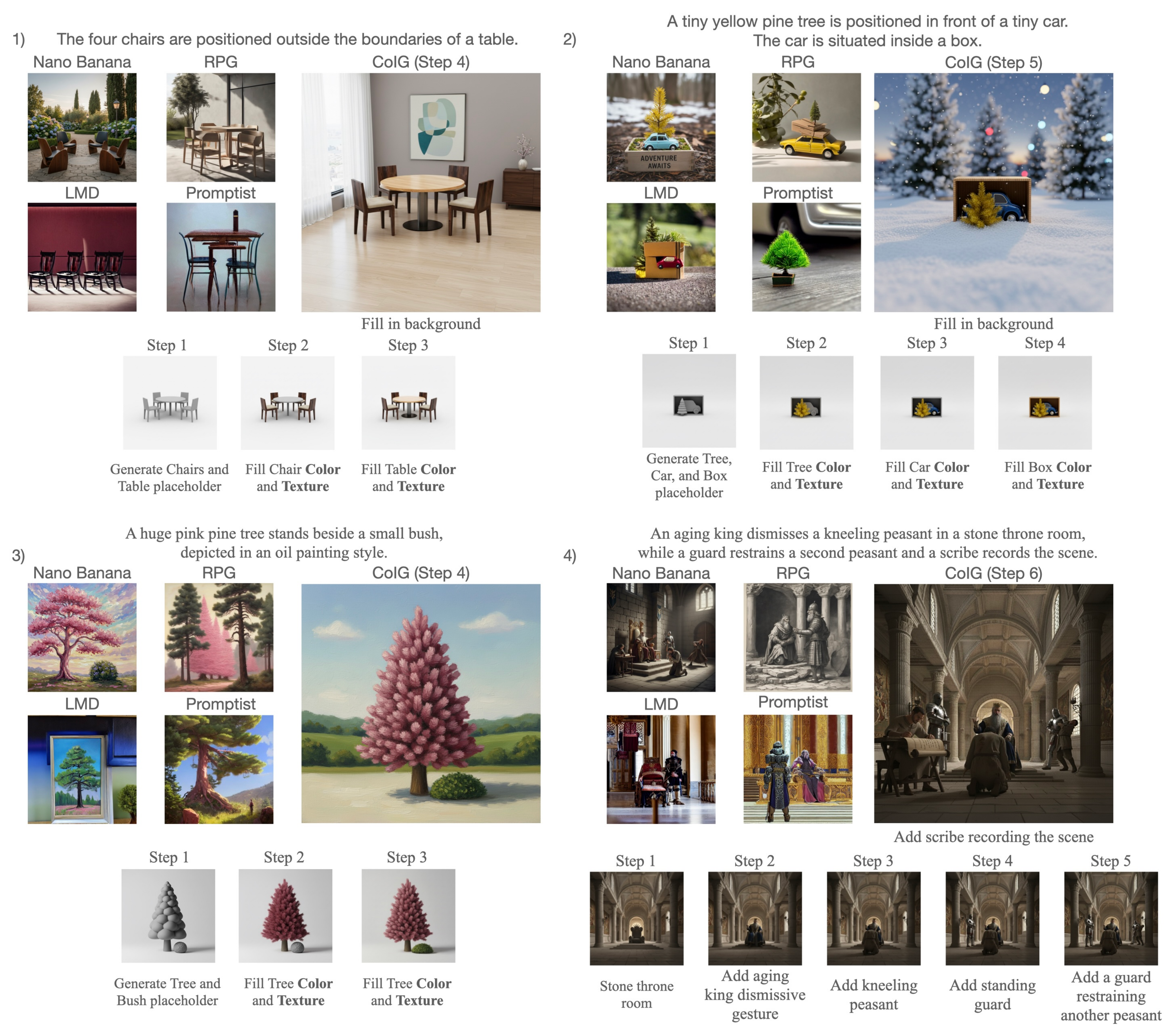}
 
 \caption{Qualitative comparison between our \textbf{CoIG} framework and four baselines—\textbf{Nano Banana}~\cite{fortin2025introducing}, \textbf{RPG}~\cite{yang2024mastering}, \textbf{LMD}~\cite{lian2023llm}, and \textbf{Promptist}~\cite{hao2023optimizing}. Across four complex scenarios that require spatial reasoning and multi-entity coordination (1--4), the baselines frequently struggle with entity collapse and incorrect spatial arrangements. In contrast, CoIG decomposes the prompt into explicit, monitorable steps (shown at the bottom of each panel), ensuring precise object placement and correct attribute binding. }
 \vspace{-0.28cm}
 \label{fig:CoIG_Read_qual_app}
\end{figure*}

Additional qualitative results are shown in Figure~\ref{fig:CoIG_Read_qual_app}.

\section{Experimental Setup}
\label{app:exp_setup}


\noindent\textbf{Evaluation Protocol:} For all experiments requiring semantic verification (GenEval, T2I-CompBench, ConceptMix, CoIG Monitorability, and the EC Benchmark), our evaluation protocol uses an MLLM evaluator adapted from~\cite{li2025easier}. We closely adopt the methodology from that paper for the T2I-CompBench and monitorability analyses. These tasks, which involve verifying compositional correctness and evaluating reasoning transparency, utilize Gemini 2.5 Flash~\cite{comanici2025gemini} as an efficient and scalable MLLM evaluator. Regarding the specific query formulation, we adapt our strategy to the dataset structure. For \textbf{GenEval} and \textbf{T2I-CompBench}, we employ a targeted QA template (e.g., ``Is the \{object\} present? Is it \{count\} in count and \{color\} in color?''), where the attribute slots are dynamically substituted with \textit{shape} or \textit{texture} depending on the specific prompt constraints. In contrast, for \textbf{ConceptMix}, we adhere to the benchmark's standardized protocol by utilizing the specific question set provided by the dataset authors. For the EC Benchmark analysis, which requires more nuanced reasoning to specifically detect complex attribute-binding failures like entity collapse, we apply a modified version of their method and utilize the more powerful Gemini 2.5 Pro~\cite{comanici2025gemini}. 

To specifically quantify \textit{Entity Collapse}, we employ a specialized ``visual census'' prompting strategy. Unlike standard boolean verification, this protocol requires the MLLM evaluator to act as a rigorous ``quality auditor.'' The model is instructed to scan the image and strictly enumerate every visible entity, assigning a unique identifier (e.g., $P_1, P_2$) to each. Crucially, the prompt enforces a \textbf{visual evidence constraint}: the evaluator must list only attributes and interactions that are unambiguously visible, ignoring any details mentioned in the text prompt that are not rendered in the image. The model outputs a structured JSON object mapping specific attributes and interactions to these unique IDs. This granular extraction allows us to algorithmically detect collapse by verifying if the number of distinct, identified entities matches the requested count ($N_{detected} = N_{requested}$) and ensuring that interaction pairs are correctly bound to separate individuals.

\noindent\textbf{Benchmarks.} We conduct our evaluations across four primary benchmarks.

\begin{figure*}[h]
 \centering
 
 \includegraphics[width=\textwidth]{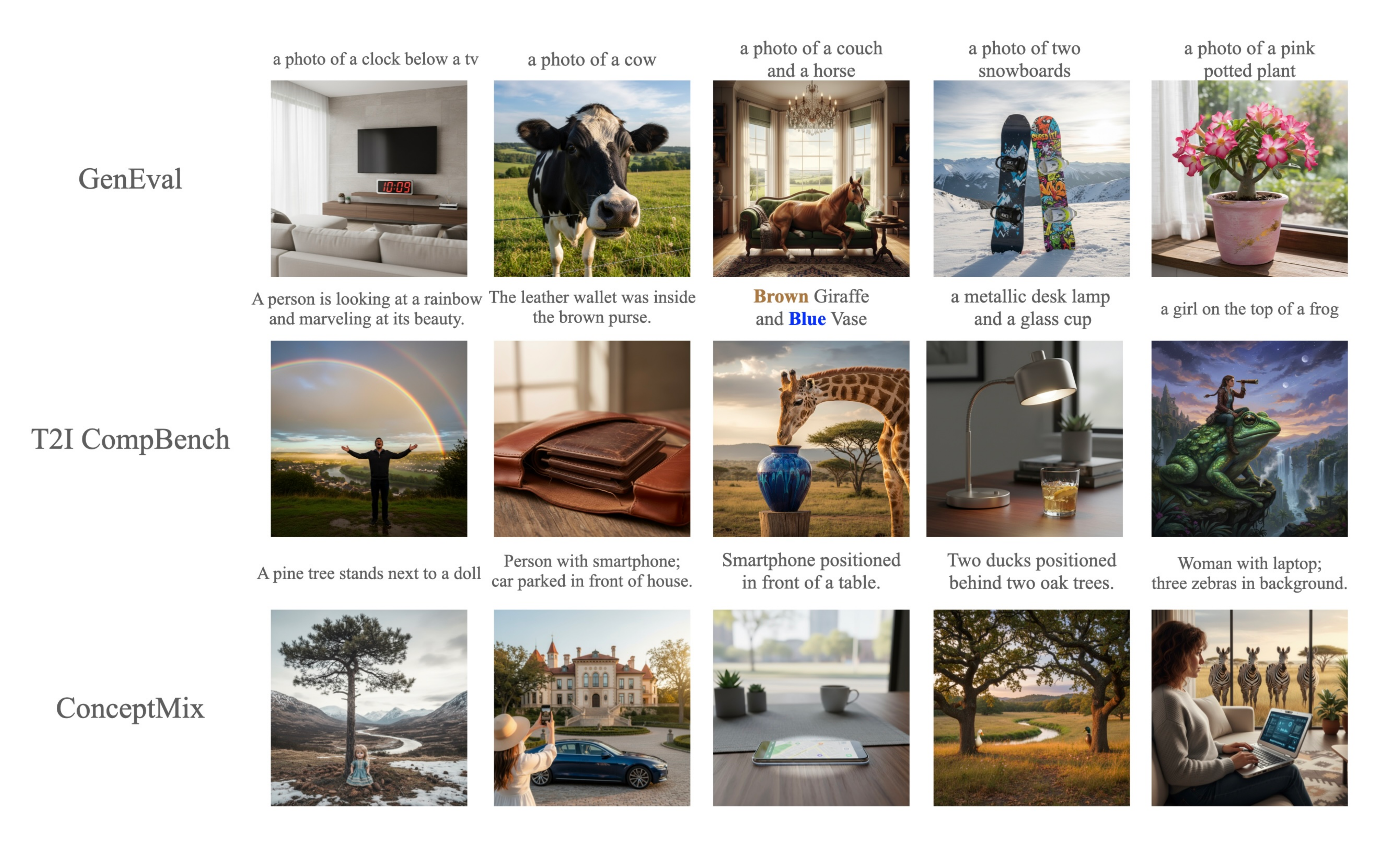}
\caption{
    \textbf{Visualizations of representative prompts from the evaluation benchmarks.} 
    We display example prompts from \textbf{GenEval} (top), \textbf{T2I-CompBench} (middle), and \textbf{ConceptMix} (bottom). 
    To illustrate the visual content and complexity described by these datasets, we generate the corresponding images using \textbf{Nano Banana}~\cite{fortin2025introducing}.
  }
 \vspace{-0.28cm}
 \label{fig:dataset_exp_app}
\end{figure*}

\noindent\textit{GenEval, T2I-CompBench, ConceptMix}~\cite{huang2023t2i}: We utilize this standard benchmark for two purposes. First, we use them for our main compositionality comparison against the Gemini baseline. These three datasets are widely used to evaluate a model's compositional abilities, each with a unique focus: GenEval provides an object-focused evaluation of properties like object color, position, and count; T2I-CompBench similarly offers structured prompts categorized into groups like attribute binding, object relationships, and complex compositions; and ConceptMix features a scalable method that automatically generates prompts by combining an increasing number of visual concepts, such as style, attribute binding, and size. Visualizations of example prompts from each dataset are shown in Figure~\ref{fig:dataset_exp_app}.
We use the T2I-CompBench benchmark's prompts as the basis for our monitorability evaluations; we assess CoIG Readability using the generated intermediate images, and we evaluate Causal Relevance by perturbing the step prompts by swapping an object's color to verify that the change is reflected in the final image.

\noindent\textit{Entity Collapse (EC) Benchmark.} We use our novel 300-prompt benchmark, as introduced in Section~\ref{sec:ec_benchmark}, to perform a targeted evaluation of the ``entity collapse'' failure mode, comparing CoIG directly against the baseline model.

\noindent\textbf{Prompts for Compositional Strategy Planner.}
To implement the Compositional Strategy Planner (CSP), we utilize a fixed system prompt that directs an off-the-shelf LLM to function as a ``systematic prompt architect.'' This prompt transforms the user's raw input into a strictly ordered sequence of semantically isolated generation steps. The decomposition process is governed by the following core algorithmic rules:

\begin{enumerate}
    \item \textbf{Foundational Anchoring (Step 1):} The generation process begins by establishing the global scene context. Depending on the complexity of the prompt, the CSP defines the first step as either a \textit{foundational sketch} (to lock the spatial layout and perspective) or a \textit{background generation} step. This ensures that the compositional structure or environmental context is fixed immediately, providing a stable canvas for subsequent additions.
    
    \item \textbf{Sequential Semantic Isolation:} Following the foundational step, the CSP decomposes the remaining prompt into a sequence where \textbf{exactly one semantic entity is processed per step}. By isolating each object, the framework ensures that the generative model focuses its attention on a single target at a time. This granular approach prevents attribute leakage and semantic confusion, allowing complex scenes to be constructed piece-by-piece.
    
    \item \textbf{Entity Persistence and Immutable Locking:} To ensure consistency and prevent ``entity collapse,'' the system prompt enforces a strict \textit{non-destructive editing policy}. Once an object has been generated or refined in a previous step, it is designated as ``locked.'' The prompt explicitly instructs the model to preserve these previously established regions, strictly forbidding any alterations to their shape, position, or appearance while new objects are being integrated. This guarantees that progress is cumulative and that earlier generations are not degraded by later steps.
    
    \item \textbf{Dual-Context Prompt Structure:} Each generated step is structured into two distinct components to maintain alignment between the global objective and local modifications:
    \begin{itemize}
        \item \textbf{Final Goal (Context):} The complete, original user caption is repeated at every step to serve as a global conditioning anchor.
        \item \textbf{This Step's Action:} A specific, imperative instruction focused solely on the current operation (e.g., ``Fill in the car with red color and metallic texture''), ensuring the generation trajectory remains precise.
    \end{itemize}
\end{enumerate}

\section{The Entity Collapse (EC) Benchmark Detail}
\label{app:ec_benchmark_detail}
To systematically evaluate robustness to entity collapse, we developed the \textit{EC Benchmark} (to be released with this paper), which includes 300 prompts procedurally generated by sampling from predefined categorical vocabularies representing \texttt{Jobs} (e.g., ``airman''), \texttt{Attributes} (e.g., ``white hair''), and \texttt{Interactions} (e.g., ``glaring at''). Each prompt is constructed by sampling one \texttt{Job} category (to enforce inter-entity similarity), four distinct \texttt{Attributes}, and two \texttt{Interactions}. This targeted prompt structure (e.g., ``Four airmen are in a room. The first is smiling and talking to the second, who is holding a bottle...'') is specifically designed to induce and measure the ``entity collapse'' failure mode. This benchmark is employed in Section~\ref{sec:experiments} to conduct a quantitative comparison of our proposed CoIG framework against baseline ``black box'' generative models.

\section{Generalizability Across Different LLM and Image Backbones}
\label{app:generalizability}

\begin{figure*}[!ht]
 
 \centering
 
 \includegraphics[width=\textwidth]{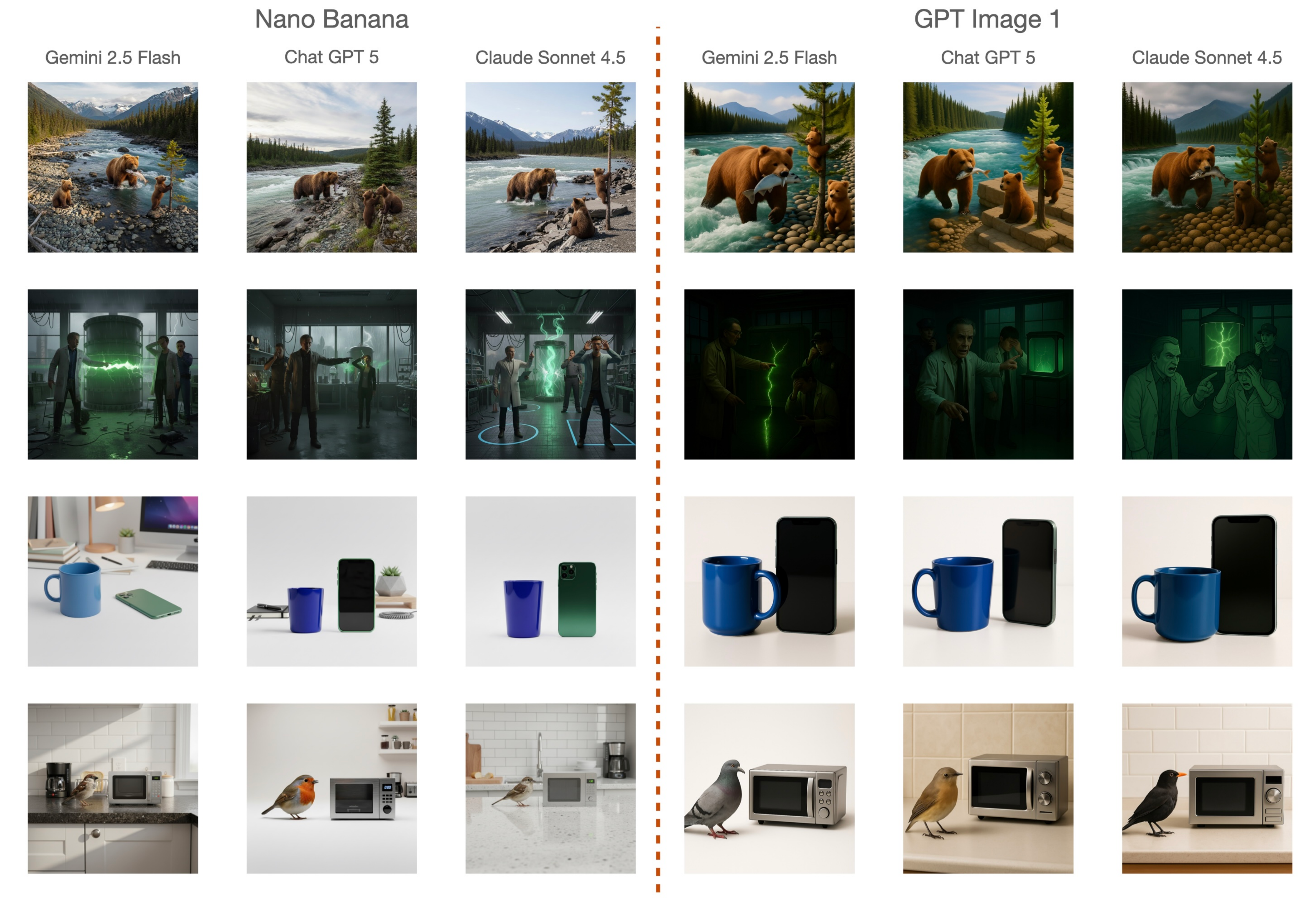}
 
 \caption{
    \textbf{Qualitative demonstration of architectural versatility.} We present final generation outputs from our CoIG framework using different backbone combinations. The figure is organized into two main sections: the \textbf{first three columns} utilize the Nano Banana image model~\cite{fortin2025introducing}, while the \textbf{last three columns} utilize the GPT Image 1 model~\cite{openai_image_gpt}. Within each section, we compare the performance of three driving LLMs: Gemini 2.5 Flash~\cite{comanici2025gemini}, Chat GPT 5~\cite{openai_gpt5}, and Claude Sonnet 4.5~\cite{claude_sonnet}. Each row represents a unique text prompt. The consistent high-quality results across this grid demonstrate that our method effectively generalizes across different model architectures.
  }
 \label{fig:ablation_app} 
\end{figure*}

A key strength of our CoIG framework is its model-agnostic architecture. The framework is designed as a versatile pipeline that can be seamlessly integrated with various Large Language Models (LLMs) and image generation architectures. To demonstrate this flexibility, we apply CoIG using three distinct LLMs (Gemini 2.5 Flash~\cite{comanici2025gemini}, Chat GPT 5\cite{openai_gpt5}, and Claude Sonnet 4.5\cite{claude_sonnet}) combined with two different image models (Nano Banana~\cite{fortin2025introducing} and GPT Image 1\cite{openai_image_gpt}). As illustrated in Figure~\ref{fig:ablation_app}, the framework consistently produces high-quality final images across all combinations. This indicates that CoIG is robust to the choice of underlying backbone models, maintaining performance stability regardless of the specific components used.

\section{Additional Qualitative Results on Readability}
\label{app:readbility_more}

\begin{figure*}[h]
 \centering
 
 \includegraphics[width=\textwidth]{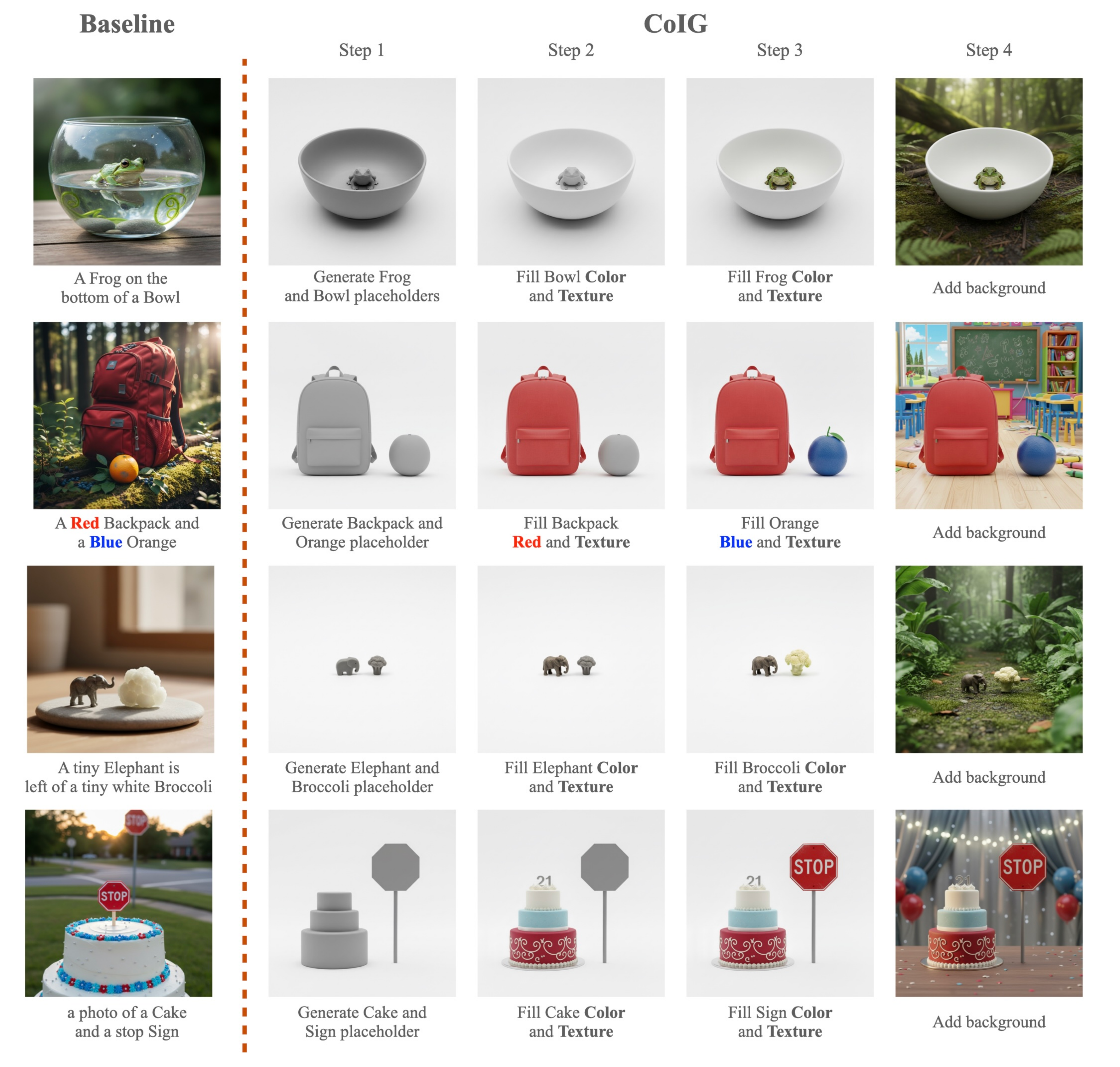}
 
 \caption{
    \textbf{Visualizing the monitorable advantage of CoIG.} 
    In contrast to the black box Baseline, which often suffers from \textbf{attribute leakage} or \textbf{semantic bias} (e.g., failing to bind ``Blue'' to the orange in row 2, or defaulting to a standard ``orange'' color), CoIG's stepwise process ensures precise attribute binding. 
    By visually separating geometry generation (Step 1), specific object coloring (Steps 2--3), and background synthesis (Step 4), our framework strictly adheres to prompt constraints and allows users to transparently monitor the generation trajectory.
  }
 \label{fig:readability_qual_app}
 
\end{figure*}

Additional qualitative results are shown in Figure~\ref{fig:readability_qual_app}.

\section{Additional Qualitative Results on Causal Relevance}
\label{app:qual_causal} 

\begin{figure*}[!ht]
 
 \centering
 
 \includegraphics[width=0.7\textwidth]{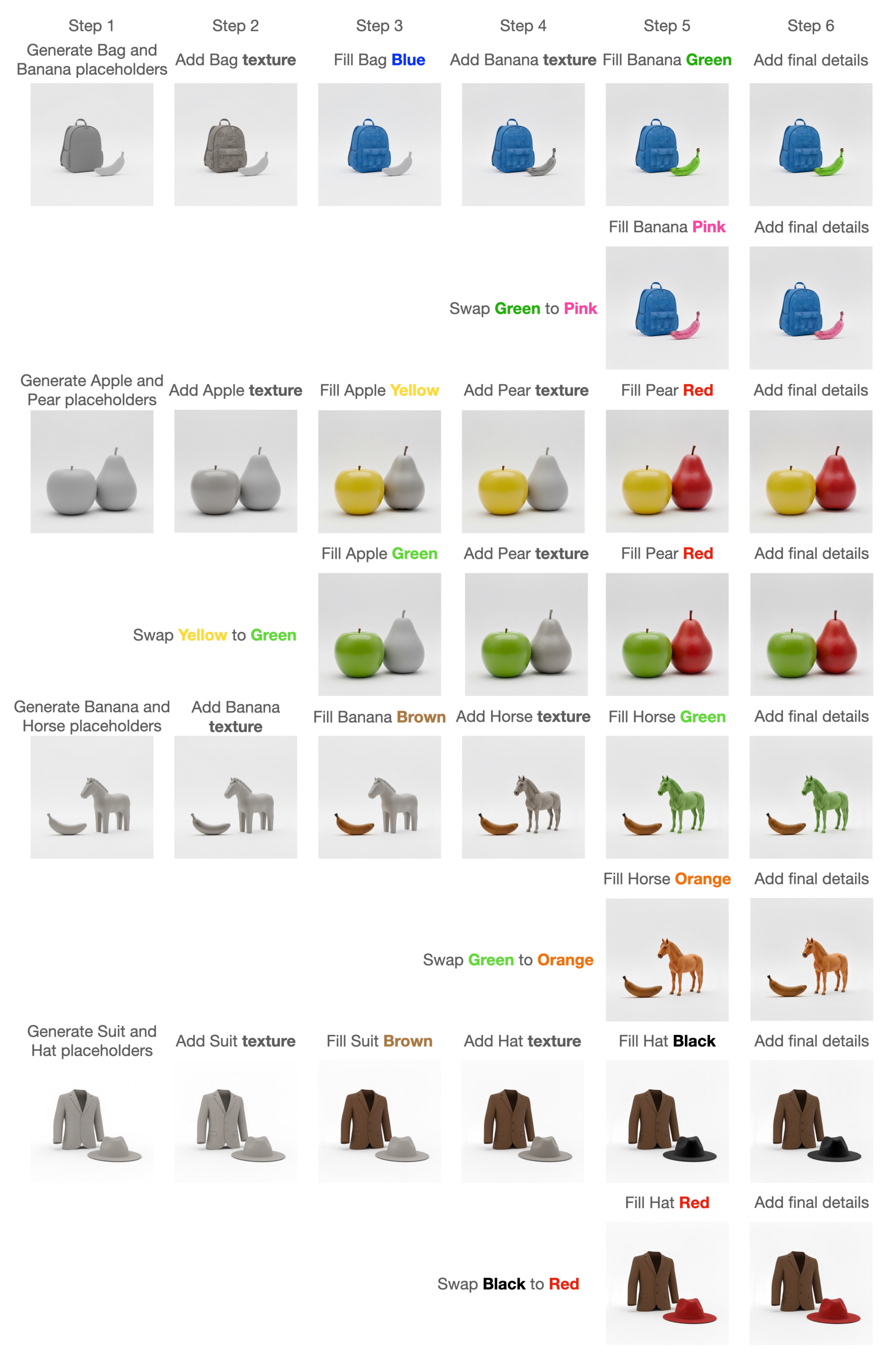}
 
 \caption{Qualitative demonstration of Causal Relevance. The odd rows (1, 3, 5, and 7) illustrate the final images generated from four distinct original CoIG sequences. The even rows (2, 4, 6, and 8) display the final images resulting from single perturbations applied during intermediate steps within those sequences. For instance, Row 2 shows the banana color changed to pink, Row 4 shows the apple changed to green, Row 6 shows the horse changed to orange, and Row 8 shows the hat changed to red. The consistent preservation of these targeted changes confirms the strong causal link between intermediate steps and the final output.}
 \label{fig:causal_qualitative_app} 
\end{figure*}

Additional qualitative results are shown in Figure~\ref{fig:causal_qualitative_app}.

\section{Additional Qualitative Results on Entity Collapse}
\label{app:entity_qual2}

\begin{figure*}[!ht]
 \centering
 
 \includegraphics[width=\textwidth]{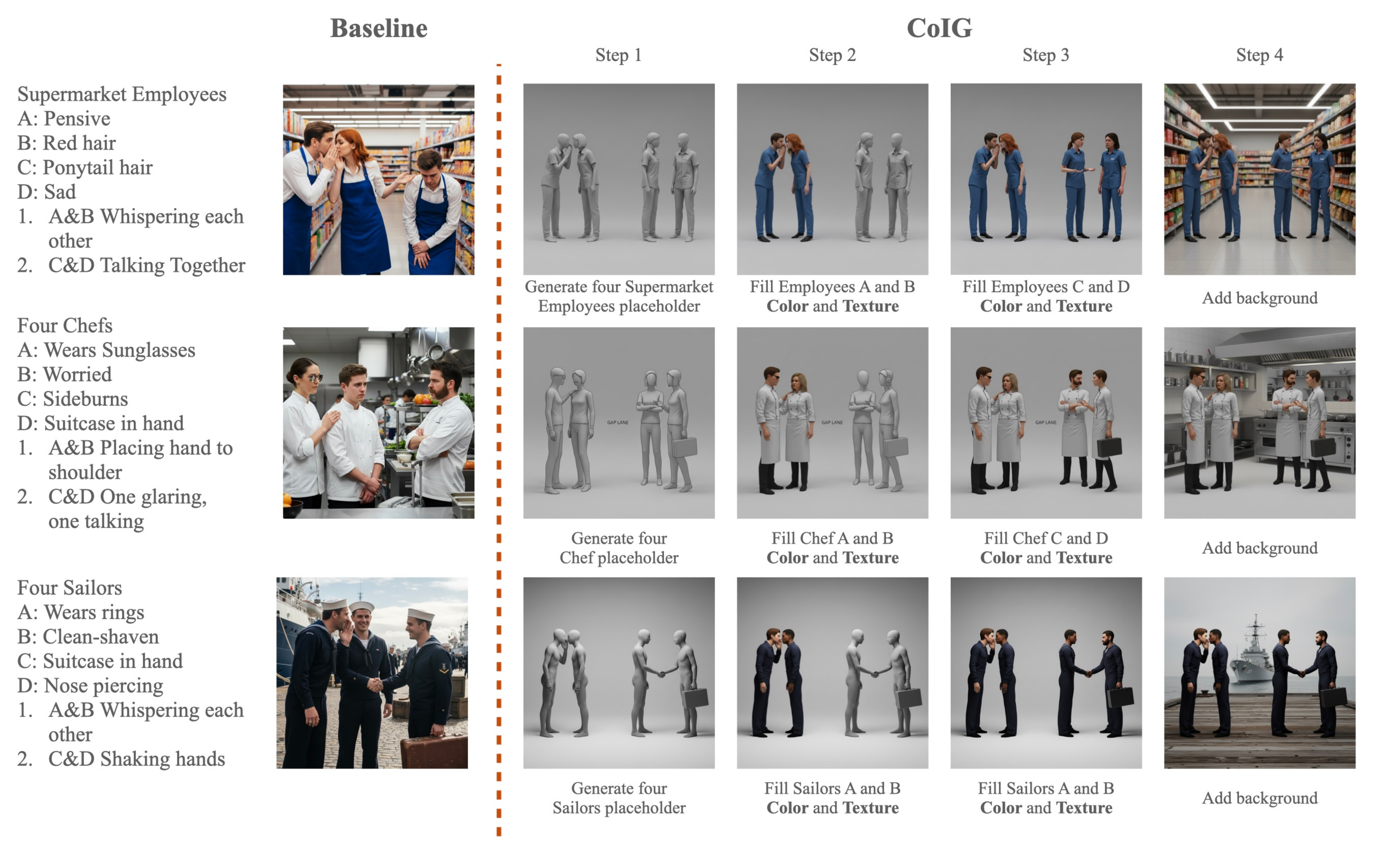}
 \caption{
    A qualitative comparison illustrating how CoIG mitigates \textbf{entity collapse}. 
    The baseline model (center) exhibits severe counting and spatial failures across all examples. 
    In Rows 1, 2, and 3, despite the prompt explicitly requesting \textbf{four} distinct individuals (Supermarket Employees, Chefs, and Sailors), the baseline consistently generates only \textbf{three} entities, effectively merging two of the subjects. 
    Consequently, the requested paired interactions (e.g., "Whispering" vs. "Shaking hands" in Row 3) become conflated or physically impossible. 
    In contrast, our proposed CoIG framework (right) successfully decomposes the task, utilizing a step-by-step generation process to strictly enforce the creation of all four distinct entities and faithfully render their specific, paired interactions.
  }
 
 \label{fig:entity_qual_app}
\end{figure*}

Additional qualitative results are shown in Figure~\ref{fig:entity_qual_app}.

\end{document}